\documentclass[twoside]{article}

\usepackage[accepted]{aistats2019}
\usepackage[round]{natbib}

\usepackage{amsfonts}
\usepackage{amssymb}
\usepackage{amsmath}
\usepackage{amsthm}

\usepackage{bm}
\usepackage{cancel}
\usepackage{bbold}
\usepackage{bm}
\usepackage{slashed}
\usepackage{color}
\usepackage{algorithm}
\usepackage{algpseudocode}
\usepackage{tikz}
\usepackage{hyperref}
\hypersetup{colorlinks=true, allcolors=blue}

\newcommand{\qxpsi}{q(\mathbf{x}|\bfpsi)}
\newcommand{\bftheta}{{\bm \theta}}
\newcommand{\bfpsi}{{\bm \psi}}
\newcommand{\bfphi}{{\bm \phi}}

\newcommand{\bfx}{\mathbf{x}}
\newcommand{\bfz}{\mathbf{z}}

\begin{document}

%
\runningtitle{Adversarial Variational Optimization}

%

\twocolumn[

\aistatstitle{Adversarial Variational Optimization \\of Non-Differentiable Simulators}


\aistatsauthor{ Gilles Louppe \And Joeri Hermans \And  Kyle Cranmer }
\aistatsaddress{ University of Li\`ege \And  University of Li\`ege \And New York University }

]

\begin{abstract}
Complex computer simulators are increasingly used across fields
of science as generative models tying parameters of an underlying theory to
experimental observations. Inference in this setup is often difficult, as
simulators rarely admit a tractable density or likelihood function. We introduce
Adversarial Variational Optimization (AVO), a likelihood-free inference
algorithm for fitting a non-differentiable generative model incorporating ideas
from generative adversarial networks, variational optimization and empirical Bayes. We adapt the
training procedure of generative adversarial networks by replacing the
differentiable generative network with a domain-specific simulator. We solve the
resulting non-differentiable minimax problem by minimizing variational upper
bounds of the two adversarial objectives.
Effectively, the procedure results in
learning a proposal distribution over simulator parameters, such that
the JS divergence between the marginal distribution of the synthetic data and
the empirical distribution of observed data is minimized.
We evaluate and compare the method with simulators producing both
discrete and continuous data.
\end{abstract}


\section{Introduction}

In many fields of science such as particle physics, epidemiology  or
population genetics, computer simulators are used to describe complex data
generation processes. These simulators relate observations $\bfx$ to the
parameters $\bftheta$ of an underlying theory or mechanistic model. In most
cases, these simulators are specified as procedural implementations of forward,
stochastic processes involving latent variables $\bfz$. Rarely do these
simulators admit a tractable density, or likelihood, $p(\bfx | \bftheta)$. The
prevalence and significance of this problem has motivated an active research
effort in so-called \textit{likelihood-free inference} algorithms such as
Approximate Bayesian Computation (ABC) and density estimation-by-comparison
algorithms~\citep{beaumont2002approximate, marjoram2003markov,
sisson2007sequential, sisson2011likelihood, marin2012approximate,
cranmer2015approximating}.

In parallel, with the introduction of variational
auto-encoders~\citep{DBLP:journals/corr/KingmaW13} and generative adversarial
networks~\citep{goodfellow2014generative}, there has been a vibrant research
program around implicit generative models based on neural
networks~\citep{2016arXiv161003483M}.  While some of these models
also do not admit a tractable density, they are all differentiable by construction.
In addition, generative models based on neural networks are highly parameterized and the model
parameters have no obvious interpretation. In contrast, scientific simulators
can be thought of as highly regularized generative models as they typically have
relatively few parameters and they are endowed with some level of
interpretation. In this setting, inference on the model parameters $\bftheta$ is
often of more interest than the latent variables $\bfz$.

In this work, we introduce Adversarial Variational Optimization (AVO), a likelihood-free inference algorithm
for non-differentiable, implicit generative models.
We adapt the adversarial
training procedure of generative adversarial
networks by replacing the implicit generative
network with a domain-based scientific simulator, and solve the resulting
non-differentiable minimax problem by minimizing variational upper
bounds of the adversarial
objectives.
The objective of the algorithm is to match the marginal distribution of
the synthetic data to the empirical distribution of observations.


\section{Problem statement}
\label{sec:problem}

We consider a family of parameterized densities $p(\mathbf{x}|\bftheta)$
defined implicitly through the simulation of a stochastic generative process,
where $\mathbf{x} \in \mathbb{R}^d$ is the data and $\bftheta$ are the
parameters of interest. The simulation may involve some complicated latent
process
where $\bfz \in {\cal Z}$ is a latent variable providing an external
source of randomness.
Unlike implicit generative models defined by neural networks, we do not assume
$\bfz$ to be a fixed-size vector with a simple density. Instead, the
dimension of $\bfz$ and the nature of its components (uniform, normal,
discrete, continuous, etc.) are inherited from the control flow of the
simulation code and may depend on $\bftheta$ in some intricate way. Moreover,
the dimension of $\bfz$ may be much larger than the dimension of
$\bfx$.

We assume that the stochastic generative process that defines $p(\mathbf{x}|\bftheta)$ is
specified through a non-differentiable deterministic function $g(\cdot; \bftheta) : {\cal Z} \to
\mathbb{R}^d$. Operationally, 
\begin{equation}\label{eqn:p_theta}
    \mathbf{x} \sim p(\mathbf{x}|\bftheta) \triangleq \bfz \sim p(\bfz|\bftheta), \mathbf{x} = g(\bfz; \bftheta)
\end{equation}
such that the density $p(\mathbf{x}|\bftheta)$ can be
written as
\begin{equation}\label{eqn:p_x_sim}
    p(\mathbf{x}|\bftheta) = \int_{\{\bfz:g(\bfz;\bftheta) = \bfx \}} p(\bfz|\bftheta) \mu(d\bfz),
\end{equation}
where $\mu$ is a probability measure.

Given some observed data $\{ \mathbf{x}_i | i=1, \dots, N \}$ drawn from the
(unknown) true distribution $p_r(\mathbf{x})$, our goal is to estimate the parameters
$\bftheta^*$ that minimize some divergence or some distance $\rho$ between $p_r(\mathbf{x})$ and
the implicit model $p(\mathbf{x}|\bftheta)$. That is,
\begin{equation}
    \bftheta^* = \arg \min_{\bftheta} \rho(p_r(\mathbf{x}), p(\mathbf{x}|\bftheta)).
\end{equation}


\section{Background}

\subsection{Generative adversarial networks}
\label{sec:gans}

Generative adversarial networks (GANs) were first proposed by
\cite{goodfellow2014generative} as a way to build an implicit generative model
capable of producing samples from random noise $\bfz$. The core principle of GANs is to pit a generative model $g(\cdot; \bftheta)$  against an adversarial
classifier $d(\cdot; \bfphi):\mathbb{R}^d \to [0,1]$ that has for antagonistic objective to recognize real data $\mathbf{x}$
from generated data $\tilde{\mathbf{x}} = g(\bfz; \bftheta)$. Both models $g$ and $d$
are trained simultaneously, in such a way that $g$ learns to fool
its adversary $d$ (which happens when $g$ produces samples comparable to the
observed data), while $d$ continuously adapts to changes in $g$.

In practice, the discriminator $d$ and the generator $g$ are usually trained with alternating stochastic gradient descent in order to respectively minimize
\begin{align}
    {\cal L}_d(\bfphi) =&\,\, \mathbb{E}_{\mathbf{x} \sim p_r(\mathbf{x})} \left[  -\log(d(\mathbf{x};\bfphi)) \right]\nonumber \\
            & + \mathbb{E}_{\tilde{\mathbf{x}} \sim p(\mathbf{x}|\bftheta)} \left[  -\log(1- d(\tilde{\mathbf{x}};\bfphi)) \right]\label{eqn:loss_g}\\
    {\cal L}_g(\bftheta) =&\,\, \mathbb{E}_{\tilde{\mathbf{x}} \sim p(\mathbf{x}|\bftheta)} \left[  \log(1 - d(\tilde{\mathbf{x}};\bfphi)) \right]\label{eqn:loss_d},
\end{align}
where ${\cal L}_d$ corresponds to the binary cross-entropy between true and synthetic data and ${\cal L}_g$ is the negative of ${\cal L}_d$ restricted to synthetic data.

When $d$ is
trained to optimality before each (infinitesimally small) parameter update of the generator, it can
be shown that the original adversarial learning procedure of \cite{goodfellow2014generative} amounts to minimizing
the Jensen-Shannon divergence $\text{JSD}$ between the distributions $p_r(\mathbf{x})$ and $p(\mathbf{x}|\bftheta)$.
Of course this assumption is never met in practice and it is often observed that the GAN alternating optimization procedure does not lead to convergence.
As a result, recent research has focused on finding better training algorithms \citep[e.g.,][]{salimans2016improved,2016arXiv161102163M,2017arXiv170107875A,2017arXiv170400028G,roth2017stabilizing} for GANs, as well as gaining a better theoretical understanding of the training dynamics~\citep[e.g.,][]{2017arXiv170104862A,mescheder2017numerics,nagarajan2017gradient}. In this work, we follow \cite{mescheder2018training} and adapt the GAN training procedure by adding a regularization term
\begin{equation}
    R_1(\bfphi) = \mathbb{E}_{\mathbf{x} \sim p_r(\mathbf{x})} \left[ || \nabla_\bfphi d(\mathbf{x};\bfphi) ||^2 \right]
\end{equation}
to the loss ${\cal L}_d$ of the discriminator. Under suitable assumptions, this regularization term guarantees the (local) convergence of the training procedure, while keeping the original GAN algorithm otherwise unchanged.

\subsection{Variational optimization}

Variational optimization~\citep{2012arXiv1212.4507S,staines2013optimization} and evolution strategies~\citep{2011arXiv1106.4487W} are general
optimization techniques that can be used to form a differentiable bound
on the optima of a non-differentiable function. Given a function $f$ to minimize,
these techniques are based on the observation that
\begin{equation}
    \min_{\bftheta} f(\bftheta) \leq \mathbb{E}_{\bftheta \sim q(\bftheta|\bfpsi)} [f(\bftheta)] = U(\bfpsi),
\end{equation}
where $q(\bftheta|\bfpsi)$ is a proposal distribution with parameters $\bfpsi$ over input values $\bftheta$.
That is, the minimum of a set of function values is always less than or equal
to any of their average. Provided that the proposal distribution is flexible enough, the parameters $\bfpsi$
can be updated to place its mass arbitrarily tight around the optimum $\bftheta^* = \min_{\bftheta \in \Theta} f(\bftheta)$.

Under mild restrictions outlined by \cite{2012arXiv1212.4507S}, the bound
$U(\bfpsi)$ is differentiable with respect to $\bfpsi$, and using the log-likelihood
trick its gradient can be rewritten as:
\begin{align}\label{eqn:approx-grad}
    \nabla_\bfpsi U(\bfpsi) &= \nabla_\bfpsi \mathbb{E}_{\bftheta \sim q(\bftheta|\bfpsi)} [f(\bftheta)] \nonumber \\
    &= \nabla_\bfpsi \int q(\bftheta|\bfpsi) f(\bftheta) d\bftheta \nonumber \\
   &= \int \nabla_\bfpsi q(\bftheta|\bfpsi)  f(\bftheta) d\bftheta \nonumber \\
   &= \int q(\bftheta|\bfpsi) \nabla_\bfpsi \log q(\bftheta|\bfpsi) f(\bftheta)  d\bftheta \nonumber \\
    &= \mathbb{E}_{\bftheta \sim q(\bftheta|\bfpsi)} [\nabla_\bfpsi \log q(\bftheta|\bfpsi) f(\bftheta) ]
\end{align}
Effectively, this means that provided that the score function $\nabla_\bfpsi \log
q(\bftheta|\bfpsi)$ of the proposal is known and that one can evaluate
$f(\mathbf{\bftheta})$ for any $\bftheta$, then one can construct empirical
estimates of Eqn.~\ref{eqn:approx-grad}, which can in turn be used to minimize
$U(\bfpsi)$ with stochastic gradient descent (or a variant thereof, robust to noise
and parameter scaling).

In reinforcement learning, Eqn.~\ref{eqn:approx-grad} similarly appears in the context of policy gradients, where $f(\bftheta)$ corresponds to a reward signal for the action $\bftheta$ and $q(\bftheta|\bfpsi)$ corresponds to a policy $\pi_\bfpsi$ that we aim to optimize. In this context, empirical estimates of  Eqn.~\ref{eqn:approx-grad} are better known as REINFORCE estimates~\citep{williams1992simple}.


\section{Adversarial variational optimization}


\subsection{Algorithm}

\begin{figure*}
    \begin{minipage}{\linewidth}
    \begin{algorithm}[H]
    \caption{Adversarial variational optimization (AVO).}

    \begin{tabular}{ l l }
        {\it Inputs:} & Observed data $\{ \mathbf{x}_i \sim p_r(\mathbf{x}) \}_{i=1}^N$, simulator $g$. \\
        {\it Outputs:} & Proposal distribution $q(\bftheta|\bfpsi)$, such that $\qxpsi \approx p_r(\mathbf{x})$. \\
        {\it Hyper-parameters:} & The number $k$ of training iterations of the discriminator $d$ (default: $k=1$), \\
                                & The size $M$ of a mini-batch (default: $M=32$),  \\
                                & The $R_1$ regularization coefficient $\lambda$ (default: $\lambda=10$), \\
                                & The entropy penalty coefficient $\gamma$ (default: $\gamma=0$).\\
                                & The baseline strategy $b$ in REINFORCE estimates (default: Eqn.~\ref{eqn:baseline}).
    \end{tabular}


    \label{alg:avo}
    \begin{algorithmic}[1]
        \State{$q(\bftheta|\bfpsi) \leftarrow \text{prior on $\bftheta$ (with differentiable and known density)}$}
        \While{$\bfpsi$ has not converged}
            \For{$i=1$ to $k$} \Comment{Update $d$}
                \State{Sample true data $\{\mathbf{x}_m \sim p_r(\mathbf{x}), y_m=1\}_{m=1}^{M/2}$.}
                \State{Sample synthetic data $\{\bftheta_m \sim q(\bftheta|\bfpsi), \bfz_m \sim p(\bfz|\bftheta_m), \tilde{\mathbf{x}}_m = g(\bfz_m; \bftheta_m), y_m=0\}_{m=M/2+1}^{M}$.}
                \State{$\nabla_\bfphi U_d \leftarrow  \frac{1}{M} \sum_{m=1}^M \nabla_\bfphi \left[ -y_m \log(d(\mathbf{x}_m;\bfphi)) - (1-y_m)\log(1-d(\mathbf{x}_m;\bfphi)) \right]$ }
                \State{$\nabla_\bfphi R_1 \leftarrow \frac{1}{M/2} \sum_{m=1}^{M/2} \nabla_\bfphi \left[ || \nabla_\bfphi d(\mathbf{x}_m;\bfphi) ||^2 \right] $}
                \State{$\bfphi \leftarrow \textsc{RmsProp}(\nabla_\bfphi U_d + \lambda \nabla_\bfphi R_1 )$}
            \EndFor
            \State{Sample synthetic data $\{ \bftheta_m \sim q(\bftheta|\bfpsi),  \bfz_m \sim p(\bfz|\bftheta_m), \tilde{\mathbf{x}}_m = g(\bfz_m; \bftheta_m) \}_{m=1}^M$.}  \Comment{Update $q(\bftheta|\bfpsi)$}
            \State{$\nabla_\bfpsi U_g \leftarrow \tfrac{1}{M} \sum_{m=1}^M \left[ \nabla_\bfpsi \log q(\bftheta_m|\bfpsi) (\log(1-d(\tilde{\mathbf{x}}_m;\bfphi)) - b) \right]$}
            \State{$\nabla_\bfpsi H \leftarrow \tfrac{1}{M} \sum_{m=1}^M  \nabla_\bfpsi \left[- q(\bftheta_m|\bfpsi) \log q(\bftheta_m|\bfpsi) \right]$}
            \State{$\bfpsi \leftarrow \textsc{RmsProp}(\nabla_\bfpsi U_g + \gamma \nabla_\bfpsi H)$}
        \EndWhile
    \end{algorithmic}
    \end{algorithm}
    \end{minipage}
\end{figure*}

The alternating stochastic gradient descent on ${\cal L}_d$ and ${\cal L}_g$ in
GANs (Section~\ref{sec:gans}) implicitly assumes that the generator $g$ is a differentiable function. In
the setting where we are interested in estimating the parameters of a
fixed non-differentiable simulator (Section~\ref{sec:problem}) --
as opposed to learning the generative model itself --
gradients $\nabla_\bftheta g$ either do not exist or are not accessible. As a
result, gradients $\nabla_\bftheta {\cal L}_g$ cannot be constructed and the
optimization procedure cannot be carried out.

In this work, we propose to rely on variational optimization to minimize ${\cal
L}_d$ and ${\cal L}_g$, thereby bypassing the non-differentiability of $g$.
We consider a proposal distribution $q(\bftheta|\bfpsi)$ over the
parameters of the simulator $g$ and alternately minimize the variational upper bounds
\begin{align}
U_d(\bfphi) &= \mathbb{E}_{\bftheta \sim q(\bftheta|\bfpsi)} [ {\cal L}_d(\bfphi) ] \label{eqn:vo-ud} \\
U_g(\bfpsi) &= \mathbb{E}_{\bftheta \sim q(\bftheta|\bfpsi)} [ {\cal L}_g(\bftheta) ] \label{eqn:vo-ug}
\end{align} respectively over $\bfphi$ and $\bfpsi$.
The discriminator $d$ is therefore no longer pit against a single generator $g$, but instead against a hierarchical family of generators induced by the proposal distribution.

When updating the discriminator parameters
$\bfphi$, unbiased estimates of $\nabla_\bfphi U_d$ can be obtained by
directly evaluating the gradient of $U_d$ over mini-batches of real and
synthetic data. When updating the proposal parameters
$\bfpsi$, $\nabla_\bfpsi U_g$ can be estimated as described in the previous section with $f(\bftheta) = {\cal L}_g(\bftheta)$.
That is,
\begin{equation}\label{eqn:grad-ug-approx}
\nabla_\bfpsi U_g = \mathbb{E}_{\substack{\bftheta \sim q(\bftheta|\bfpsi), \\ \tilde{\bfx} \sim p(\bfx | \bftheta)}}  [\nabla_\bfpsi \log q(\bftheta|\bfpsi) \log(1-d( \tilde{\bfx} ;\bfphi))],
\end{equation}
which we can approximate with mini-batches of synthetic data.

While the latter REINFORCE-like gradient estimator is unbiased, it is well known that it also suffers from high variance, which makes the optimization unstable and difficult.
A common remedy to this issue~\citep{williams1992simple} is to make use of the fact that
\begin{align}
    &\mathbb{E}_{\bftheta \sim q(\bftheta|\bfpsi)} [\nabla_\bfpsi \log q(\bftheta|\bfpsi) f(\bftheta) ] \nonumber \\
    &= \mathbb{E}_{\bftheta \sim q(\bftheta|\bfpsi)} [\nabla_\bfpsi \log q(\bftheta|\bfpsi) (f(\bftheta) - b) ]
\end{align}
for any constant $b$.
The choice of the baseline $b$ does not bias the gradient estimator, but it can however have an effect on its variance.
For AVO, we pick the baseline which minimizes the variance of the empirical estimates of $\nabla_\bfpsi U_g$, that is
\begin{equation}
    b=\frac{\mathbb{E} \left[ (\nabla_\bfpsi \log q(\bftheta|\bfpsi))^2 (\log(1-d(\tilde{\bfx};\bfphi))^2 \right]}{\mathbb{E} \left[ (\nabla_\bfpsi \log q(\bftheta|\bfpsi))^2 \right]}.\label{eqn:baseline}
\end{equation}

For completeness, Algorithm~\ref{alg:avo} outlines the proposed Adversarial Variational
Optimization (AVO) procedure, as built on top of GAN with $R_1$ regularization.


\subsection{Parameter Point Estimates}

The variational objectives \ref{eqn:vo-ud}-\ref{eqn:vo-ug}
effectively replace the modeled data distribution of Eqn.~\ref{eqn:p_theta} with
the parameterized marginal distribution of the generated data
\begin{equation}\label{eq:marginal_likelihood}
\qxpsi = \int  p(\mathbf{x}|\bftheta) q(\bftheta|\bfpsi) d\bftheta.
\end{equation}
We can think of $q(\bfx|\bfpsi)$ as a \textit{variational program} as described
by \cite{2016arXiv161009033R}, though more complicated than a simple
reparameterization of normally distributed noise $\bfz$ through a differentiable
function. In our case, the variational program is a
marginalized, non-differentiable  simulator.  Its density is  intractable;
nevertheless, it can generate samples for $\bfx$ whose expectations are differentiable with
respect to $\bfpsi$.
Operationally, we sample from this marginal model via
\begin{equation}\label{eqn:p_psi}
\mathbf{x} \sim \qxpsi \triangleq \bftheta \sim q(\bftheta|\bfpsi), \bfz \sim p(\bfz|\bftheta), \mathbf{x} = g(\bfz; \bftheta).
\end{equation}

We can view the optimization of $q(\bfx | \bfpsi)$ with respect to $\bfpsi$ through the lens of empirical Bayes, where the data are used to optimize a prior within the family $q(\bftheta|\bfpsi)$.
Since the GAN procedure effectively minimizes the Jensen-Shannon divergence between $p_r(\bfx)$ and $\qxpsi$, $\bfpsi^*$ corresponds with the maximum marginal likelihood estimator advocated by \cite{rubin1984}.
When the model is well specified, $\bfpsi^*$
coincides with the true data-generating parameter; however, if the model is
misspecified, the $\bfpsi^*$ is typically different from the maximum likelihood
estimator (MLE).
Thus, if the simulator $p(\bfx |
\bftheta)$ is misspecified, $q(\bftheta | \bfpsi)$ will attempt to smear it so that the marginal
model $q(\bfx | \bfpsi)$ is closer to $p_r(\bfx)$. However, if the simulator is well specified,
then $q(\bftheta | \bfpsi)$ will concentrate its mass around the true data-generating parameter.

In order to more effectively target
point estimates $\bftheta^*$, we can also augment Eqn.~\ref{eqn:vo-ug} with
an entropic regularization term $H(q(\bftheta|\bfpsi))$, such that
\begin{equation}
U_g = \mathbb{E}_{\bftheta \sim q(\bftheta|\bfpsi)} [ {\cal L}_g ] + \gamma H(q(\bftheta|\bfpsi)),
\end{equation}
where $\gamma \in \mathbb{R}^+$ is a hyper-parameter controlling the trade-off
between the generator objective and the tightness of the proposal distribution and $H$ is the Shannon entropy.
For small values of $\gamma$,
proposal distributions with large entropy are not penalized, which results
in learning a smeared variation of the original simulator. On the other hand,
for large values of $\gamma$, the procedure is encouraged to fit a proposal
distribution with low entropy, which has the effect of concentrating its density
tightly around one or a few $\bftheta$ values.

Finally, we note that very large penalties may
eventually make the optimization unstable, as the variance of $\nabla_\bfpsi \log q(\bftheta_m|\bfpsi)$
typically increases as the entropy of the proposal decreases. Depending on the proposal, it
may also be possible to always arbitrarily decrease the entropy, without necessarily producing accurate parameter estimates. In this case, properly controlling for $\gamma$ and the number of training epochs is critical.


\section{Experiments}


\subsection{Illustrative example}
\label{sec:poisson}

As a first illustrative experiment, we evaluate inference for a discrete Poisson
distribution with unknown mean $\lambda$. We artificially consider
the distribution as a parameterized simulator, from which we can only
generate data.

The discrete observed data is sampled from a Poisson with mean $\lambda^* = 7$.
Algorithm~\ref{alg:avo} is run for 3000 iterations with mini-batches of size $M=32$
and the following configuration. For the discriminator $d$, we use a 3-layer MLP with 20
hidden nodes per layer and PReLU activation units.
For estimating $\lambda^*$, we parameterize $\bftheta$ as $\log(\lambda)$ and use a univariate Gaussian proposal distribution
$q(\bftheta|\bfpsi)$ initialized with a mean of $\log(1)$ and a variance of $0.5^2$.
The $R_1$ regularization coefficient is set to
$10$, and the entropy penalty is evaluated at both $\gamma=0$ and $\gamma=0.0001$.
The learning rate of $\textsc{RmsProp}$ is set to $0.001$, both for the discriminator $d$ and the proposal $q$.

\begin{figure}
\centering
\includegraphics[width=0.48\textwidth]{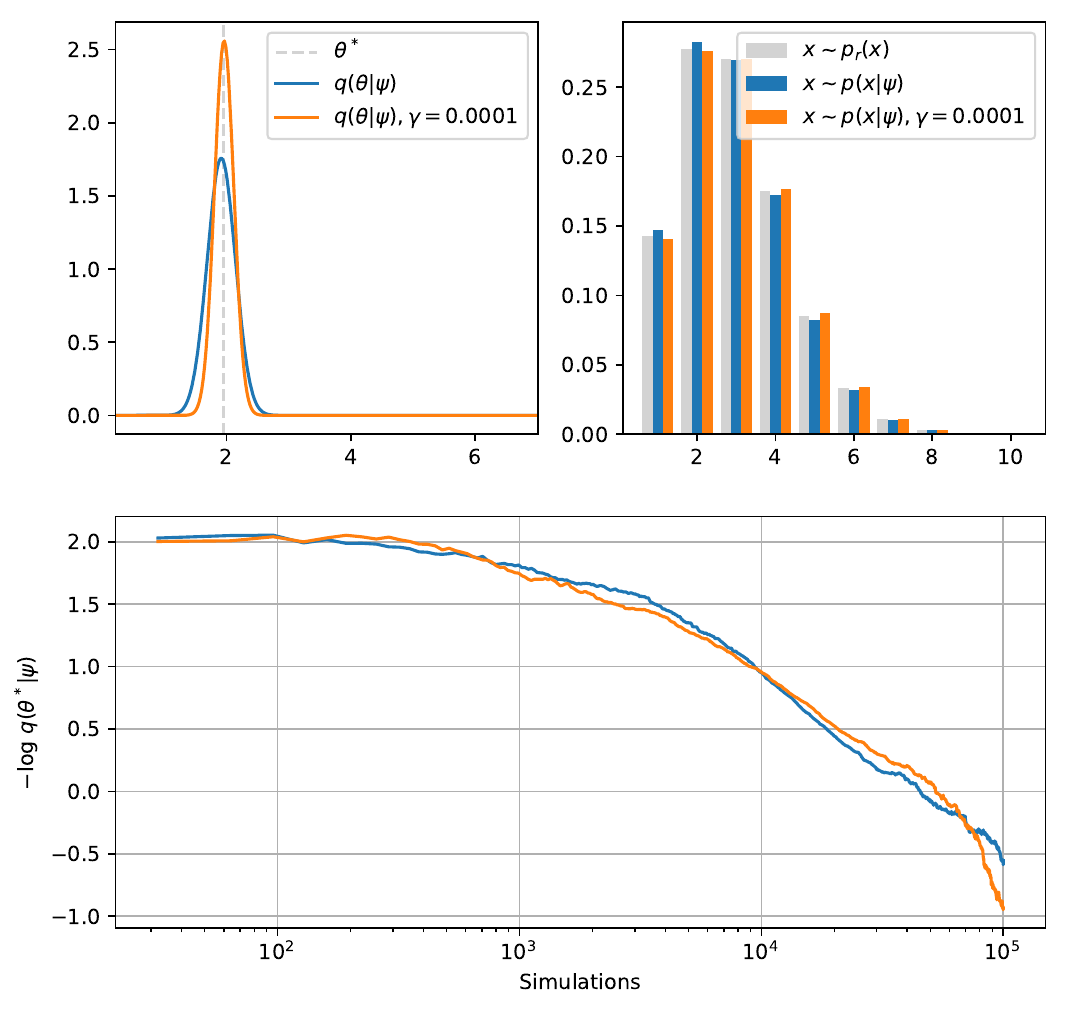}
\caption{Discrete Poisson model with unknown mean.
 ({\it Top left}) Proposal distributions $q(\bftheta|\bfpsi)$ after training. For both $\gamma=0$ and $\gamma=0.0001$, the distributions correctly concentrate their density around
            the true value $\log(\lambda^*)$. Entropic regularization ($\gamma=0.0001$) results in a tighter density.
 ({\it Top right}) Model distributions $\qxpsi$ after training. This plot shows that the resulting parameterizations of the simulator closely reproduce the true distribution.
 ({\it Bottom}) Negative log-likelihood $-q(\bftheta^*|\bfpsi)$ of the target parameters, as function of the number of simulated samples.
 }\label{fig:poisson}
\end{figure}

The top left plot in Figure~\ref{fig:poisson} illustrates the resulting proposal
distributions $q(\bftheta|\bfpsi)$ after running AVO.  For
both $\gamma=0$ and $\gamma=0.0001$, the proposal distributions correctly concentrate
their density around the true parameter value $\log(\lambda^*) = 1.94$. Under
the effect of entropic regularization,
the proposal distribution for $\gamma=0.0001$ concentrates its mass more tightly,
yielding in this case more precise inference.  The top right plot compares the
model distributions to the true distribution.  As theoretically expected from
adversarial training, we see that the resulting distributions align with
the true distribution, with in this case visually slightly better results for the penalized
model.  The bottom plot of Figure~\ref{fig:poisson} evaluates the negative log-likelihood
of the true parameters $\lambda^*$ with respect to the number of simulated samples.
For the two settings, the loss steadily decreases as the proposal converges towards
the nominal parameter value. This short example highlights that adversarial variational optimization
works despite the discreteness of the data and the lack of access to the
density $p(\bfx | \bftheta)$ or its gradient.

\begin{figure}
\centering
\includegraphics[width=0.48\textwidth]{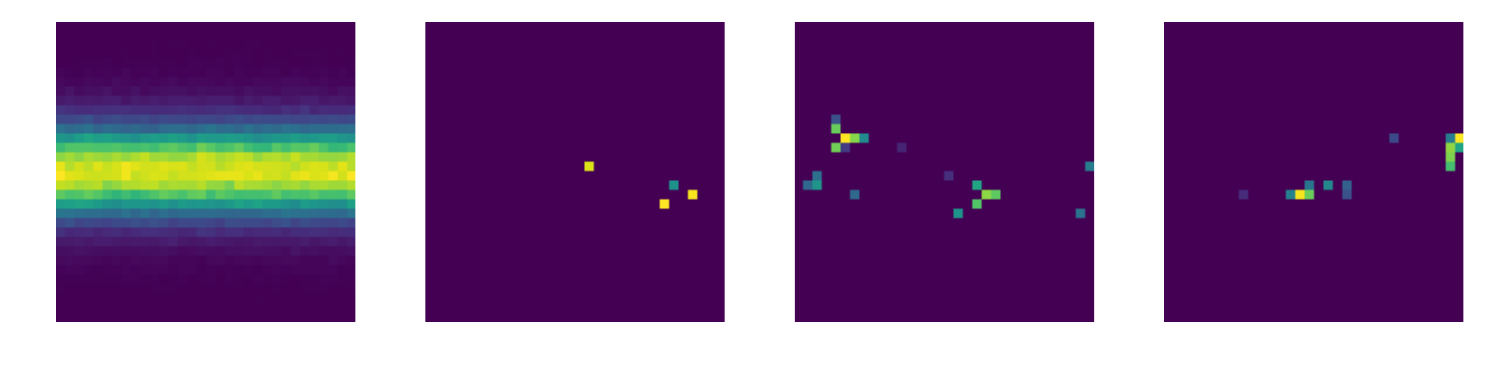}
\includegraphics[width=0.48\textwidth]{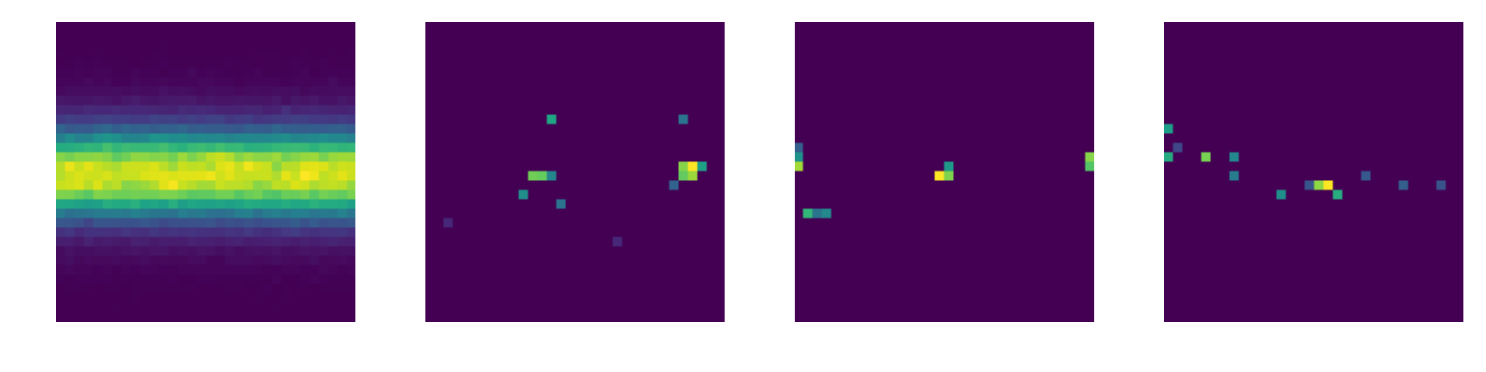}
\caption{\textsc{Pythia}-alignment: Samples. ({\it Top row}): Average detector response (approximated over 200,000 samples) for the detector offset $\bftheta=0$, along with 3 individual random samples. ({\it Bottom row}) Same but for the detector offset $\bftheta=1$.
Given the sparsity and variability of the simulated events, these plots highlight the difficulty in observing a difference between samples from one or the other parameter setting.}
\label{fig:pythia-toy-detector-responses}
\end{figure}

\subsection{High-energy particle collisions}
\label{sec:pythia-toy}

As a more challenging example, we now turn to a particle physics inference problem.
We consider the \textsc{Pythia} simulator~\citep{sjostrand2008brief} for high-energy particle collisions routinely used by physicists at the Large Hadron Collider.
We simulate electron-positron collisions  at a center-of-mass energy of 91.2 GeV, in which a $Z$ boson is produced and decays to quarks. We assume a detector that emulates a $32 \times 32$ spherical uniform grid in pseudorapidity $\eta$ and in azimuthal angle $\phi$, covering $(\eta, \phi) \in [-5,5]\times[0,2\pi]$. The detector is parameterized by an offset parameter $\bftheta$ in the z-axis relative to the beam crossing point~\citep{PythiaMill}. An offset of $\bftheta=0$ means that the sphere is centered at the collision point, while $\bftheta=1$ leads to a shift of roughly one pixel.

The inference problem we are interested in is the estimation of the offset parameter $\bftheta$ from a set of $32 \times 32$-dimensional observations. This task is representative of calibration and alignment tasks, which are critical in experimental particle physics as
they have significant impact on the accuracy of reconstruction algorithms.

The leftmost plots of Figure~\ref{fig:pythia-toy-detector-responses} show the average detector response for two distinct offsets $\bftheta=0$ and $\bftheta=1$. The remaining plots illustrate individual random samples from these respective configurations. The figures highlight the challenging difficulty of the inference problem: the difference between the average responses is barely noticeable,
while individual samples are very sparse and reflect a wide range of variability. These samples also stress the critical role of a relevant summary statistic on such high-dimensional data, which is required in  likelihood-free inference methods such as ABC.

\begin{figure}
\centering
\includegraphics[width=0.48\textwidth]{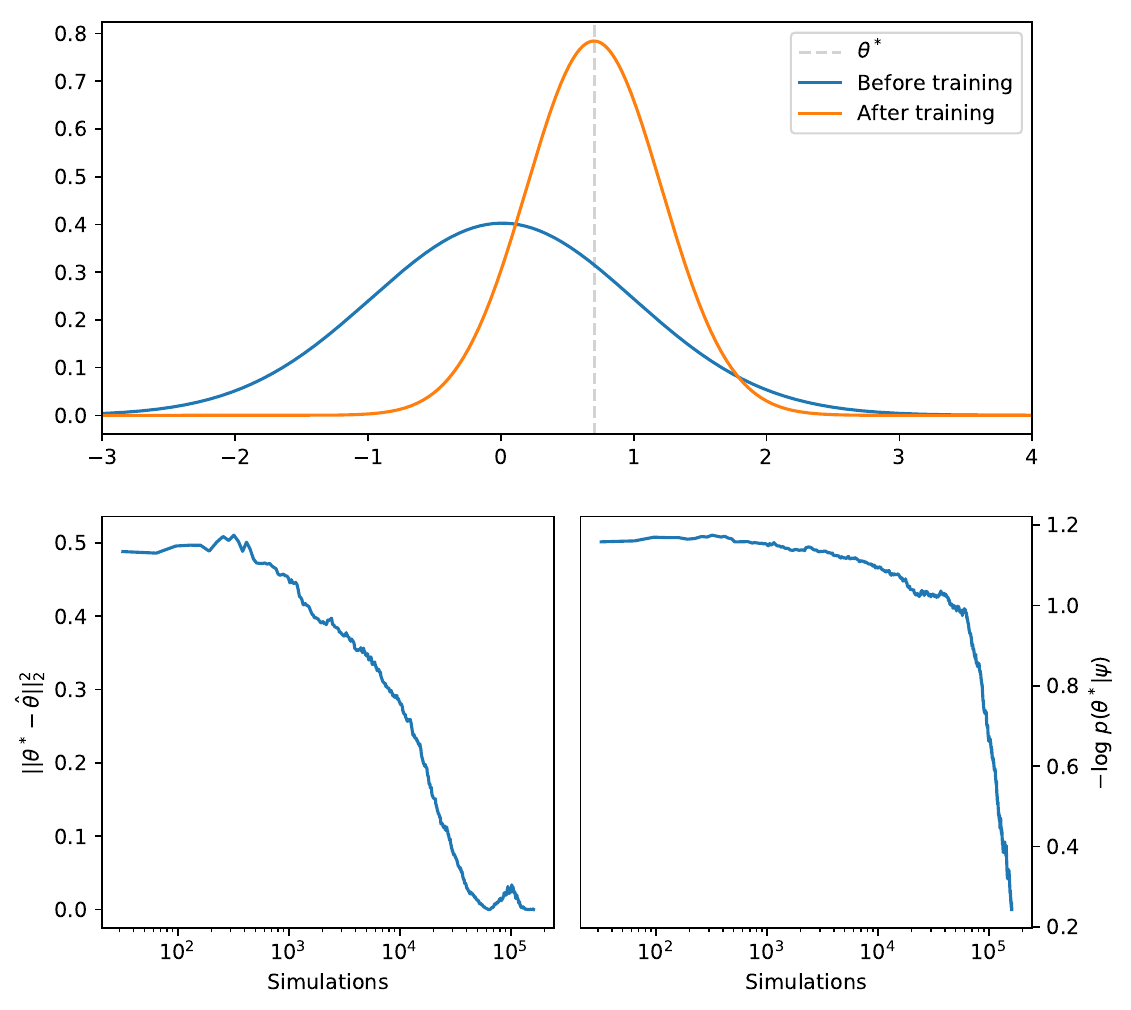}
\caption{\textsc{Pythia}-alignment: training. ({\it Top}) Proposal distribution $q(\bftheta|\bfpsi)$ before and after AVO.
({\it Bottom left}). Distance between the target parameter $\bftheta^*$ and the average parameter under the proposal, as a function of the number of simulated samples.
({\it Bottom right}). Negative log-likelihood of the target parameter value value $\bftheta^*$ under the proposal, as a function of the number of simulated samples.}
\label{fig:pythia-toy-summary}
\end{figure}

For this experiment, we consider observed data simulated at the nominal value $\bftheta^*=1$. Algorithm~\ref{alg:avo} is run for 5000 iterations with all hyper-parameters set to their default values. The discriminator $d$ is defined as a 9-layer MLP with 600 hidden nodes per layer and PReLU activations. The proposal distribution is initialized as a Gaussian with zero mean and unit variance.
As shown in the top plot of Figure~\ref{fig:pythia-toy-summary}, the proposal distribution properly centers around the target $\bftheta^*=1$ after training.
The bottom plots in the figure also illustrate the convergence of AVO as a function of the number of simulations. Despite the complexity of the $\textsc{Pythia}$ simulator, the sparsity, variability and high-dimensionality of the observations, as well as the absence of any domain knowledge, AVO properly identifies the target parameter within a reasonable number of simulations.
As suggested clearly by the bottom right plot of Figure~\ref{fig:pythia-toy-summary}, where the negative log-likelihood $-\log q(\bftheta^*|\bfpsi)$ has not yet converged,  more accurate results could certainly be obtained by running AVO for more iterations. Equivalently, we anticipate room for hyper-parameter tuning.

Finally, let us also comment on the bump around $10^5$ simulations in the left bottom plot of  Figure~\ref{fig:pythia-toy-summary}. This illustrates the particular scenario in which a temporary deviation in the mean of the proposal from the target parameter value is compensated by the variance of the proposal, which  thereby results in even lower negative log-likelihood  $-\log q(\bftheta^*|\bfpsi)$. In particular, this is confirmed by the right bottom plot of the figure, where no such bump is observed.


\subsection{Benchmarks}
\label{sec:benchmarks}

\paragraph{Methods.}
In this section, we systematically evaluate AVO on benchmark problems. We compare our algorithm against ABC-SMC~\citep{toni2009simulation} and BOLFI~\citep{gutmann2016bayesian} as baselines.
ABC-SMC is the most commonly used instance of Approximate Bayesian Computation. It makes use of importance sampling to improve efficiency.
BOLFI is a simulation-efficient likelihood-free inference algorithm that combines Bayesian optimization with a Gaussian process-based approximation of the likelihood of summary statistics of the data.
For ABC-SMC and BOLFI, we respectively use the \textsc{PyABC}~\citep{Klinger162552} and the \textsc{ELFI}~\citep{1708.00707} implementations.
All hyper-parameters are set to the default values recommended in these packages.

\paragraph{Inference tasks.} We evaluate all three methods on four inference tasks of increasing difficulty. These tasks include discrete, continuous, low-dimensional and high-dimensional observations. For each task, we evaluate the quality of inference in terms of squared error for 15 different target parameter values $\bftheta^*_{i}$, for $i=1, \dots, 15$. For each target value, we consider a data set with 100,000 observations representing $\bfx \sim p_r(\bfx)$. All methods  evaluated share the same simulation budget (160,000 samples).

\begin{itemize}
    \setlength\itemsep{0.0em}
    \item {\it Poisson.} This inference problem is the same as in Section~\ref{sec:poisson}, with $\bftheta^*_{i} \sim {\cal U}(0, 4)$. The discriminator $d$ is defined as a 3-layer MLP with PReLU activation units and 600 nodes per hidden layer.
    \item {\it \textsc{Carl}-Multidimensional.} We reproduce the inference problem originally defined in Section 4.2 of \citep{cranmer2015approximating}. The generator is parameterized by two parameters $\alpha$ and $\beta$ and produces 5-dimensional continuous data $\bfx \in \mathbb{R}^5$. For our benchmark, we consider $\alpha^*_{i} \sim {\cal U}(-2,2)$ and $\beta^*_{i} \sim {\cal U}(-2,2)$. The discriminator $d$ is defined as a 4-layer MLP with PReLU activation units and 100 nodes per hidden layer.
    \item {\it Weinberg.} We consider a simplified simulator for electron-positron collisions, as described in Appendix~\ref{sec:weinberg}. We consider $E^\text{beam}_{i} \sim {\cal U}(43, 47)$ and $G^f_{i} \sim {\cal U}(0, 2)$. The discriminator $d$ is defined as 4-layer MLP with PReLU activation units and 1000 nodes per hidden layer.
    \item {\it \textsc{Pythia}-alignment.} This inference problem is the same as in Section~\ref{sec:pythia-toy}, with $\bftheta^*_{i} \sim {\cal U}(-1.5, 1.5)$. The discriminator $d$ is defined as a 9-layer MLP with PReLU activation units and 600 nodes per hidden layer.
\end{itemize}

The summary statistics used in ABC-SMC and BOLFI are the same. For Poisson, \textsc{Carl}-Multidimensional and Weinberg, the summary statistics are defined as the Euclidean distance between (the bins of) the histogram of the observations generated at $\bftheta^*_i$ and (the bins of) the histogram of simulated data. For \textsc{Pythia}-alignment, the summary statistics is defined as the $\ell_2$ norm between the average image of the observed data and the average image of the simulated samples at the model parameter. In both methods, 128 simulation samples are generated per model parameter evaluation. The priors used are identical to the uniform priors used for generating the 15 problems $\bftheta^*_i$. For AVO, the proposal distribution is initialized as a Gaussian of zero mean and unit variance.

In contrast to some related works, we focus on the setting where we have more than one observation $\bfx$ from the data distribution $p_r(\bfx)$. For this reason, we do not consider likelihood-free benchmarks such as the M/G/1 queue model, the Lotka-Volterra population model or the Hodgkin-Huxley neuron model, which are all defined as inference problems from single observations. We anticipate that AVO is less appropriate for this use case, as the discriminator $d$ would not be expected to provide a good learning signal for fitting the simulator parameters.  The proper treatment of this scenario is left as future work.

\begin{figure}
\centering
\includegraphics[width=0.48\textwidth]{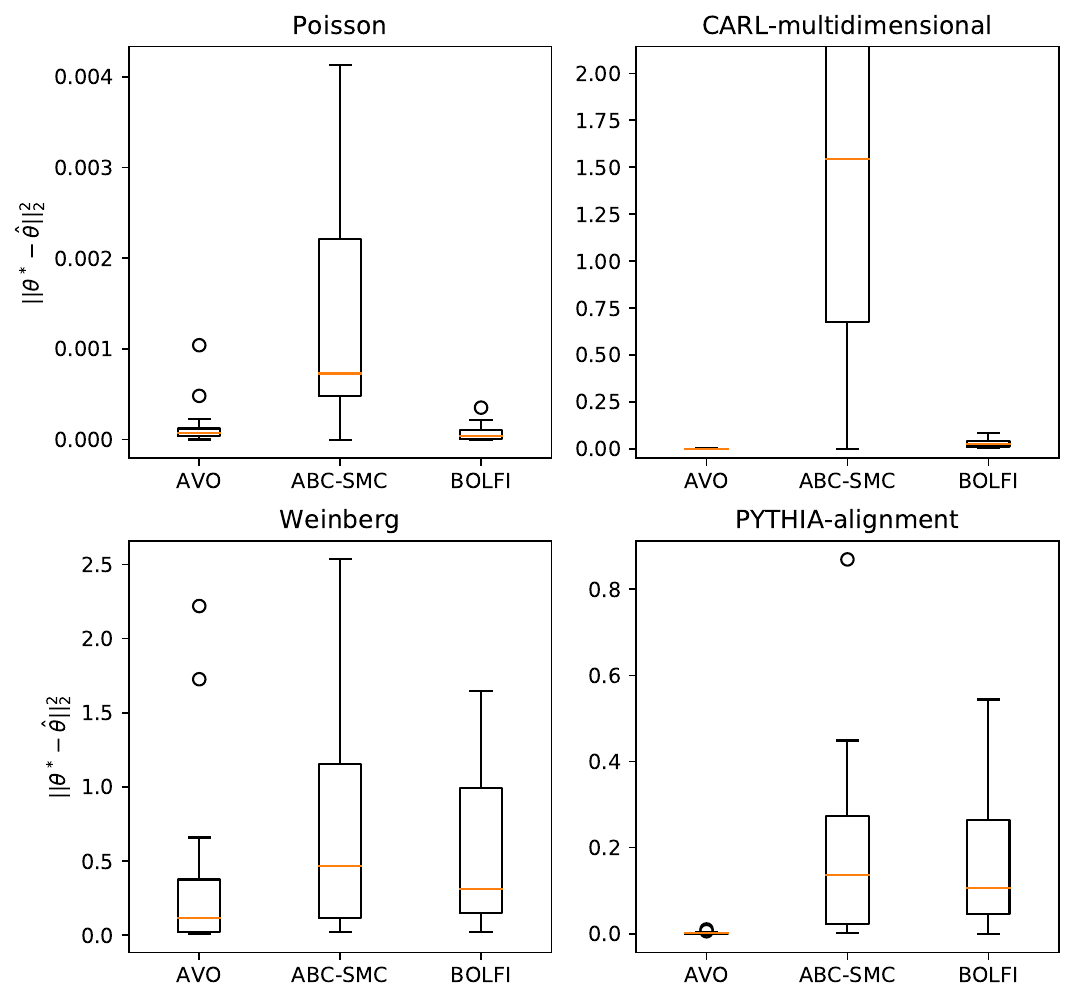}
\caption{Benchmark results comparing AVO against ABC-SMC and BOLFI.
AVO shows superior performance against these methods. This advantage stems from the fact that AVO is not limited by the sub-optimality of an ad hoc summary statistic. Instead, the discriminator in AVO can dynamically adapt to the inference problem.
 }\label{fig:result1}
\end{figure}

\paragraph{Results.}
Figure~\ref{fig:result1} summarizes our results for AVO, ABC-SMC and BOLFI on each of the four inference problems. Each plot reports as a box plot the squared distance of the best fit $\hat{\bftheta}_i$ to the nominal model parameters $\bftheta^*_i$, for $i=1, \dots, 15$. In AVO, $\hat{\bftheta}_i$ corresponds to the mode of the final proposal. For ABC-SMC and BOLFI,  $\hat{\bftheta}_i$ is the maximum a posteriori estimate under the posterior. Best fit values are comparable since we assume uniform priors.

The figure clearly indicates AVO works better on average compared to ABC-SMC and BOLFI.
We attribute this superior performance primarily to the fact that AVO is not limited by the deficiencies of a hand-crafted summary statistic. Instead, AVO benefits from a high-capacity discriminator that dynamically adapts to the inference problem and to the current proposal. This is clearly apparent for \textsc{Pythia}-alignment, where a generic summary statistic leads to a sub-optimal estimator. By contrast, because of the high-capacity discriminator $d$, AVO has no issue in guiding the proposal towards a solution, despite the high-dimensionality of the observations or the complexity of underlying generative process. Of course, ABC-SMC and BOLFI can be improved by engineering better summary statistics, but this requires a deep understanding of the problem.
While it is not illustrated here, the active learning strategy of BOLFI shows better sample efficiency than AVO, in the sense that it can often reach a good solution within a smaller simulation budget.
Finally, for the Weinberg benchmark, we observe that there is no clear winner in terms of the squared error.
This is mainly due to the an approximate degeneracy between the parameters leading to a very broad minimum and a number of solutions that fit the observed data distribution (see Appendix~\ref{sec:weinberg-results}).


\section{Related work}

This work sits at the intersection of several lines of research related to
likelihood-free inference, approximate Bayesian computation (ABC),
implicit generative models, and variational inference.
Viewed from the literature around implicit generative models based on neural networks,
the proposed method can be considered as a direct adaptation of
generative adversarial networks~\citep{goodfellow2014generative} to
non-differentiable simulators using variational optimization~\citep{2012arXiv1212.4507S}.
From the point of view of likelihood-free inference, where  non-differentiable
simulators are the norm, our contributions are threefold. First is the process
of lifting the expectation with respect to the non-differentiable simulator
$\mathbb{E}_{\tilde{\bfx} \sim p(\bfx | \bftheta)}$ to a differentiable
expectation with respect to the variational program $\mathbb{E}_{\tilde{\bfx}
\sim q(\bfx | \bfpsi)}$. Secondly, is the introduction of a novel form of
variational inference that works in a likelihood-free setting. Thirdly, AVO can be viewed as a form of empirical Bayes where the prior is
optimized based on the data.

As for many likelihood-free inference algorithms, AVO is intimately tied to a class of algorithms that can be framed as
density estimation-by-comparison, as reviewed in \citep{2016arXiv161003483M}. In most cases, these
inference algorithms are formulated as an iterative two-step process where the
model distribution is first compared to the true data distribution and then
updated to make it more comparable to the latter. Relevant work in this
direction includes those that rely on a classifier to estimate the discrepancy between the observed data and the model distributions
\citep{gutmann2012noise,cranmer2015approximating,cranmer2016experiments,2016arXiv161110242D,gutmann2017likelihood,rosca2017variational}.
Of direct relevance in the likelihood-free setup, Hamiltonian ABC~\citep{meeds2015hamiltonian}
estimates gradients with respect
to $\bftheta$ through finite differences from multiple forward passes of the
simulator with variance reduction strategies based on controlling the source of
randomness used for the latent variable $\bfz$.
Sharing similar foundational principles as AVO but developed independently, the SPIRAL algorithm~\citep{2018arXiv180401118G}
makes use of the Wasserstein GAN objective and variants of REINFORCE gradient estimates
to adversarially train an agent that synthesizes programs controlling a non-differentiable
graphics engine in order to reconstruct target images or perform unconditional generation.

Likewise, AVO closely relates to recent extensions of GANs, such as
ALI~\citep{dumoulin2016adversarially},
BiGANs~\citep{donahue2016adversarial},
$\alpha$-GAN~\citep{rosca2017variational}, AVB~\citep{DBLP:journals/corr/MeschederNG17}, and the \texttt{PC-Adv}
algorithm of  \citep{2017arXiv170208235H}, which
add an inference network to the generative model.  Each of these assume a
tractable density $p(\bfx|\bftheta)$ that is differentiable with respect to
$\bftheta$, which is  not satisfied in the likelihood-free setting. Our lifting
of the non-differentiable simulator $p(\bfx|\bftheta)$ to the variational
program $q(\bfx | \bfpsi)$ provides the ability to differentiate expectations
with respect to $\bfpsi$ as in Eqn~\ref{eqn:approx-grad}; however, the density
$q(\bfx | \bfpsi)$ is still intractable. Moreover, we do not attempt to define a
recognition model  $q(\bfz, \bftheta|\bfpsi)$ as the latent space $\mathcal{Z}$
of many real-world simulators is complicated and not amenable to a neural
recognition model.

This work has also many connections to work on variational inference, in which
the goal is to optimize the recognition model $q(\bfz, \bftheta|\bfpsi)$ so that
it is close to the true posterior $p(\bfz, \bftheta |\mathbf{x})$. There have
been efforts to extend variational inference to intractable likelihoods;
however, many require restrictive assumptions.  In \citep{tran2017variational},
the authors consider Variational Bayes with an Intractable Likelihood (VBIL). In
that approach ``the only requirement is that the intractable likelihood can be
estimated unbiasedly.'' In the case of simulators, they propose to use the
ABC-likelihood with an $\epsilon$-kernel. The ABC likelihood is only unbiased as
$\epsilon \to 0$,   thus this method inherits the drawbacks of the
ABC-likelihood including the choice of summary statistics and the inefficiency
in evaluating the ABC likelihood for high-dimensional data and small $\epsilon$.
More recently, \citep{2017arXiv170208896T} adapted variational inference to
hierarchical implicit models defined on simulators. In this work, the authors
step around the intractable likelihoods by reformulating the optimization of the
ELBO in terms of a neural and differentiable approximation $r$ of the log-likelihood
ratio $\log \tfrac{p}{q}$, thereby effectively using the same core principle as used
in GANs~\citep{2016arXiv161003483M}. With a similar objective, \citep{2017arXiv171203353M}
adapt variational inference to a non-differentiable cardiac simulator by maximizing
the ELBO using Bayesian optimization, hence bypassing altogether the need for gradient estimates.



\section{Summary}

In this work, we develop a likelihood-free inference algorithm for
non-differentiable, implicit generative models. The algorithm combines
adversarial training  with variational optimization to minimize variational
upper bounds  on the otherwise non-differentiable adversarial objectives. The
AVO algorithm enables empirical Bayes through variational inference in the
likelihood-free setting. This approach does not incur the inefficiencies of an
ABC-like rejection sampler nor the disadvantages of likelihood-free inference algorithms that rely on ad hoc summary statistics. When the model is well-specified, the AVO algorithm
provides point estimates for
the generative model, which asymptotically corresponds to the data generating
parameters.
Experimental results highlight the good performance of AVO in comparison to the well-established ABC-SMC and BOLFI algorithms.


\appendix

\section{Weinberg benchmark}

\subsection{Simulation}
\label{sec:weinberg}

For this benchmark inference task, we consider a simplified simulator from
particle physics for electron--positron collisions resulting in muon--antimuon
pairs ($e^+e^- \rightarrow \mu^+\mu^-$). The simulator approximates the
distribution of observed measurements $\mathbf{x} = \cos(A) \in [-1,1]$, where $A$ is the
polar angle of the outgoing muon with respect  to the originally incoming
electron. Neglecting measurement uncertainty induced from the particle detectors,
this random variable is approximately distributed as
\begin{equation*}
p(\mathbf{x}|E^\text{beam}, G^f) = \frac{1}{Z} \left[ (1 + \mathbf{x}^2) + c(E^\text{beam}, G^f) \mathbf{x} \right]
\end{equation*}
where $Z$ is a known normalization constant and $c$ is an asymmetry coefficient
function. Due to the linear term in the expression, the density $p(\mathbf{x} |
E^\text{beam}, G^f)$ exhibits a so-called {\it forward-backward} asymmetry.  Its
size depends on the values of the parameters $E^\text{beam}$ (the beam energy)
and $G^f$ (the Fermi constant) through the coefficient function $c$.

A typical physics simulator for this process includes a more precise treatment of the
quantum mechanical  $e^+e^- \rightarrow \mu^+\mu^-$ scattering
using \textsc{Pythia} or \texttt{MadGraph}~\citep{Alwall:2011uj},  ionization of matter in the
detector due to the passage of the out-going $\mu^+\mu^-$ particles using
\texttt{GEANT4}~\citep{Agostinelli:2002hh}, electronic noise and other details of the sensors
that measure the ionization signal, and the deterministic algorithms that
estimate the polar angle $A$ based on the sensor readouts. The simulation of
this process is highly non-trivial as is the space of latent variables $\mathcal{Z}$.

\subsection{Results}
\label{sec:weinberg-results}

\begin{figure}
\centering
\includegraphics[width=0.48\textwidth]{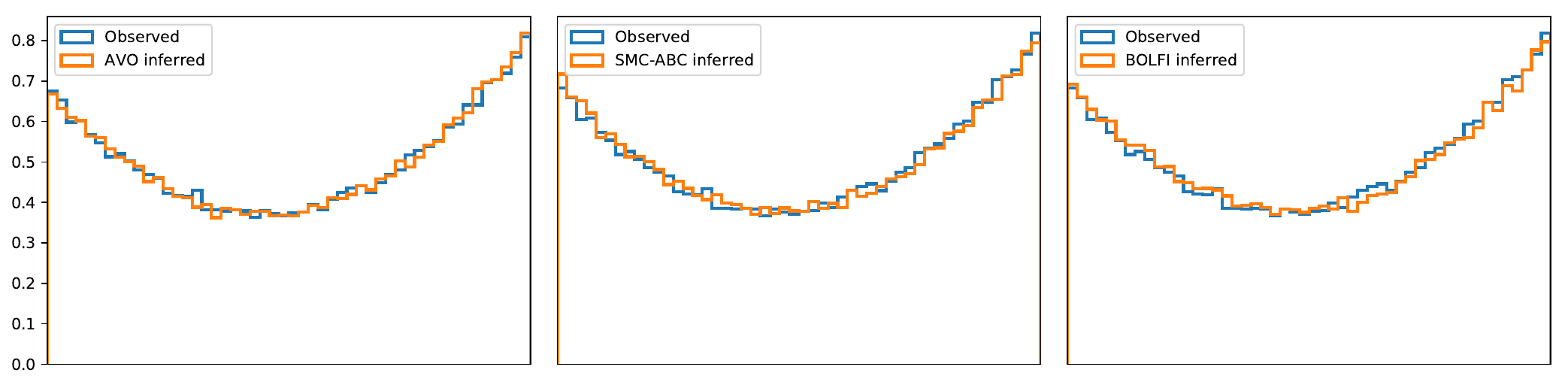}
\includegraphics[width=0.48\textwidth]{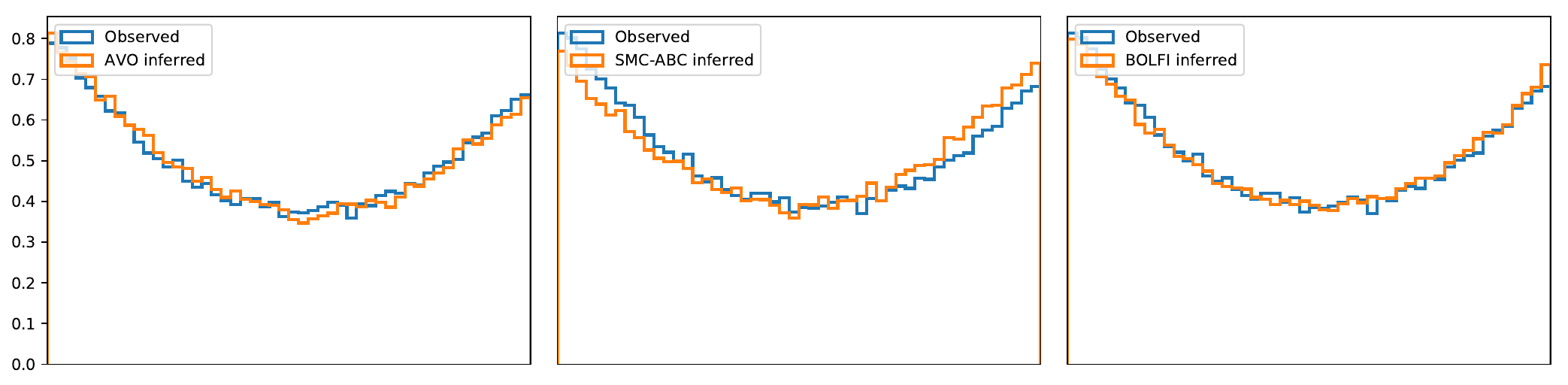}
\includegraphics[width=0.48\textwidth]{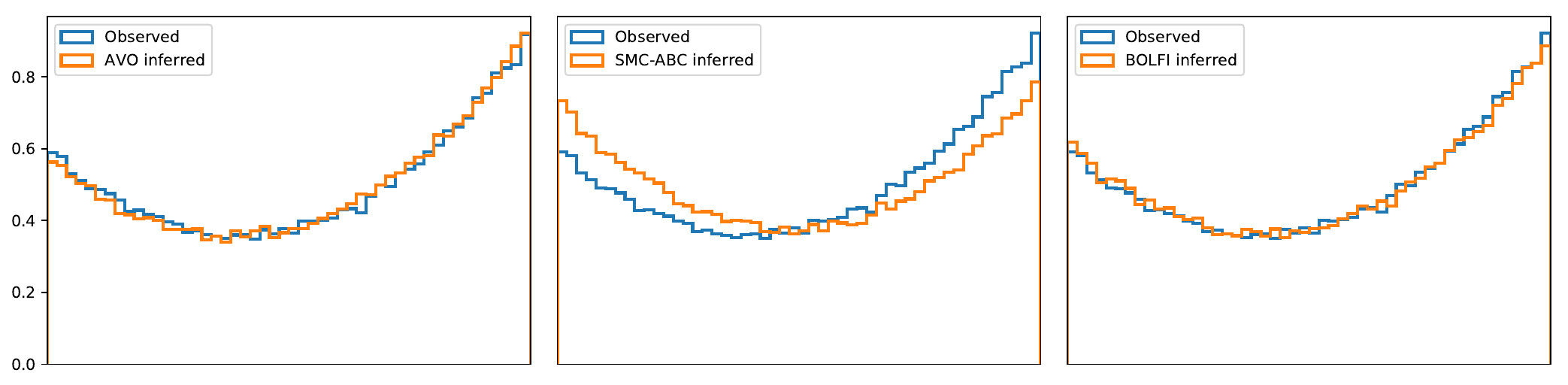}
\includegraphics[width=0.48\textwidth]{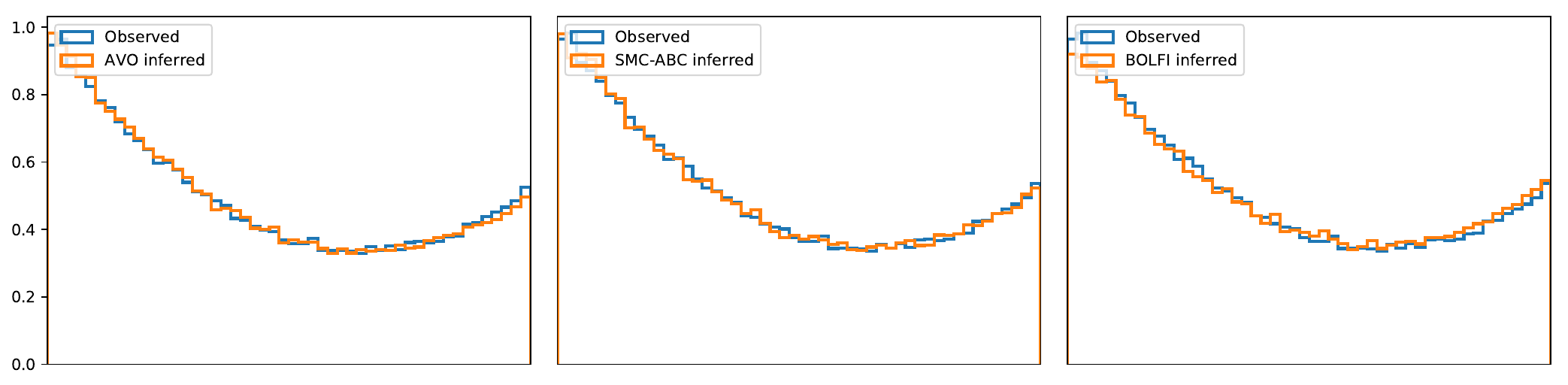}
\includegraphics[width=0.48\textwidth]{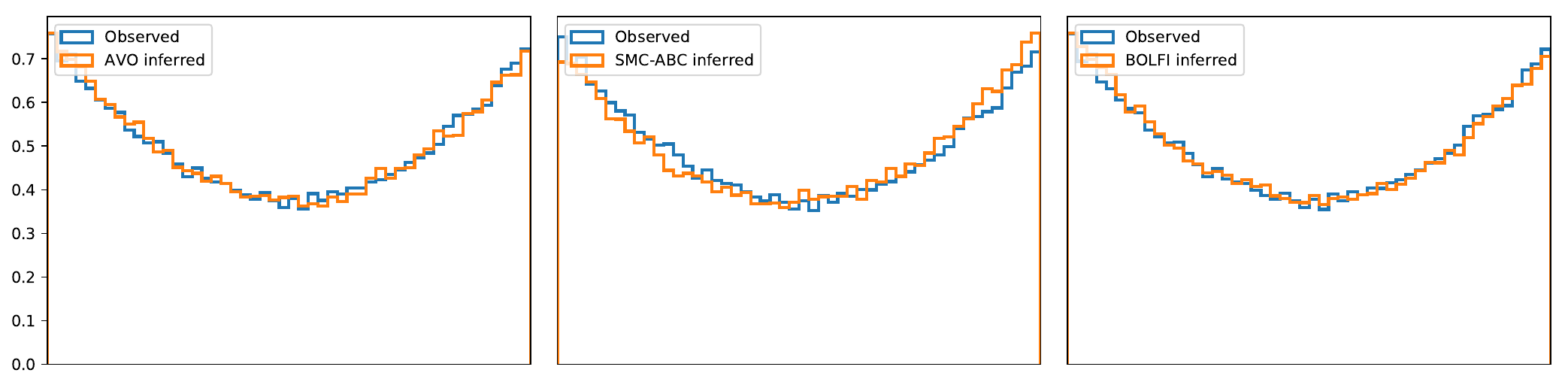}
\includegraphics[width=0.48\textwidth]{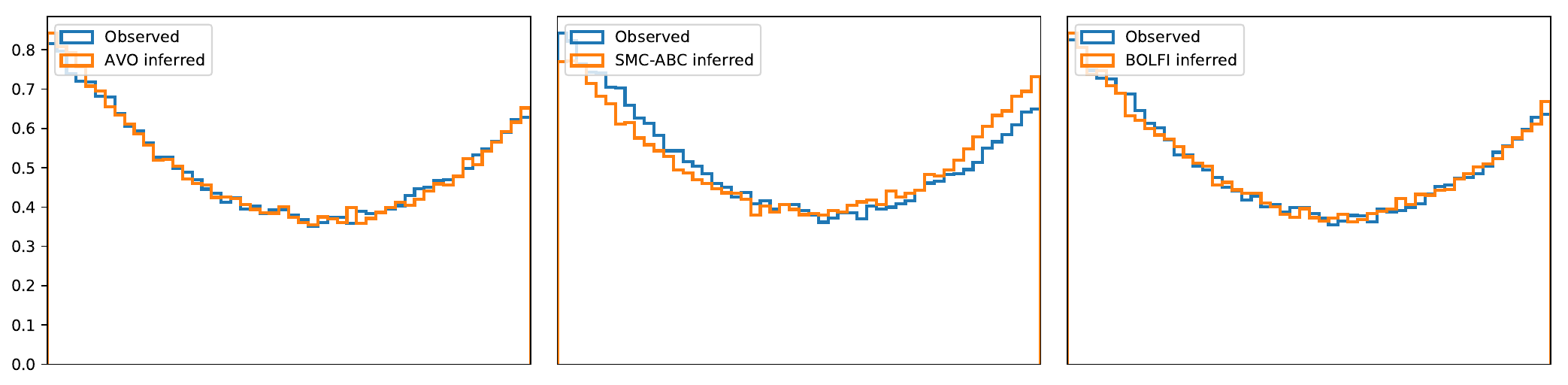}
\includegraphics[width=0.48\textwidth]{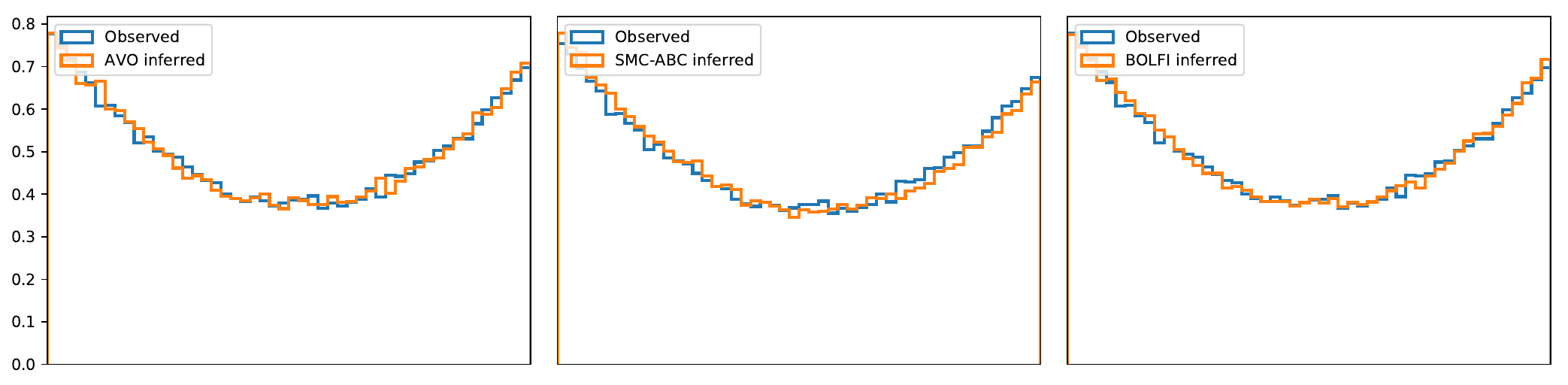}
\caption{(\emph{Left}) AVO. (\emph{Center}) SMC-ABC. (\emph{Right}) BOLFI. Despite the apparent poor performance of AVO, SMC-ABC and BOLFI in Figure~\ref{fig:result1}, all methods approximate the observed data distribution $p_r(\mathbf{x})$ for different $\bftheta^*_i$ (rows). This discrepancy is attributed to multiple minima in the Weinberg benchmark.}
\label{fig:weinberg-distributions-methods}
\end{figure}

A prominent issue with the Weinberg benchmark is the presence of a nearly degenerate direction for the likelihood in the model parameter space. This leads to a number of solutions that provide good fits to the observed data. Since Figure~\ref{fig:result1} evaluates $\vert\vert \bftheta^* - \hat{\bftheta} \vert\vert^2_2$, the presence of this broad minima significantly influences the result. To show that AVO, SMC-ABC, and BOLFI do find solutions that describe the data well, we sample $\mathbf{x} \sim p(\mathbf{x}|\hat{\bftheta})$ (inferred) and compare against $p_r(\mathbf{x})$ (observed) for several $\bftheta^*_i$, as shown in Figure~\ref{fig:weinberg-distributions-methods}. These plots demonstrate that for this benchmark, there exist many equivalent solutions that induce the observed data, even if they may be quite distant in parameter space.

\subsection*{Acknowledgments}

We would like to thank Lukas Heinrich for helpful comments regarding
the electron--positron annihilation simulation. We would also like to thank
Rajesh Ranganath and Dustin Tran for enlightening discussions and feedback.
GL and KL were both supported through NSF ACI-1450310 at the time of the research, additionally KC is
supported through PHY-1505463 and PHY-1205376. GL is recipient of the ULiège-NRB Chair on
Big Data and is thankful for the support of NRB.
JH acknowledges the financial support from F.R.S-FNRS for his FRIA PhD scholarship.

\bibliography{bibliography}

\end{document}